\documentclass[11pt,a4paper]{article}

\usepackage[colorlinks]{hyperref}
\usepackage[colorinlistoftodos]{todonotes}
\usepackage{verbatim}
\usepackage{tabularx,pbox}
\usepackage[hyperref]{acl2018}
\usepackage{times}
\usepackage{latexsym}
\usepackage{amsmath}
\usepackage{amsfonts}
\usepackage{color}
\usepackage{graphicx}
\usepackage{booktabs}

\usepackage{url}

\usepackage{subcaption}

\aclfinalcopy % Uncomment this line for the final submission
\title{A Sequence-to-Sequence Model for Semantic Role Labeling}

\author{Angel Daza and Anette Frank \\
   Leibniz ScienceCampus ``Empirical Linguistics and Computational Language Modeling''\\
   Department of Computational Linguistics\\
   Heidelberg University \\
   69120 Heidelberg, Germany\\
  {\tt \{daza,frank\}@cl.uni-heidelberg.de}
 \\}

\date{}

\begin{document}
\maketitle
\begin{abstract}
We explore a novel approach for Semantic Role Labeling (SRL) by casting it as a sequence-to-sequence process. We employ an attention-based model enriched with a copying mechanism to ensure faithful regeneration of the input sequence, while enabling interleaved generation of argument role labels. Here, we apply this model in a monolingual setting, performing PropBank SRL on English language data. The constrained sequence generation set-up enforced with the copying mechanism allows us to analyze the performance and special properties of the model on manually labeled data and benchmarking against state-of-the-art sequence labeling models.
We show that our model is able to solve the SRL argument labeling task on English data, yet further structural decoding constraints will need to be added to make the model truly competitive. Our work represents a first step towards more advanced, generative SRL labeling setups. 
\end{abstract}

\section{Introduction}
Semantic Role Labeling (SRL) is the task of assigning semantic argument structure to constituents or phrases in a sentence, to answer the question: {\it Who} did {\it what} to {\it whom}, {\it where} and {\it when}? This task is normally accomplished in two steps: first, identifying the predicate and second, labeling its arguments and the roles that they play with respect to the predicate. SRL 
has been formalized in different frameworks, the most prominent being
FrameNet \cite{Baker1998TheProject} and PropBank \cite{Palmer2005TheRoles}. In this work we focus on \textit{argument identification and labeling} using the PropBank (PB) annotation scheme.

Recent end-to-end neural models considerably improved the state-of-the-art results for SRL in English \cite{He2017DeepNext,Marcheggiani2017GCN}. In general, such models treat the problem as a supervised sequence labeling task, using deep LSTM architectures that assign a label to each token
within the sentence. 

SRL training resources for other languages are more restricted in size and thus, models suffer from sparseness problems because specific predicate-role instances occur only a handful of times in the training set. Since annotating SRL data in larger amounts is expensive, the use of a generative neural network model could be beneficial for automatically obtaining more labeled data in low-resource settings.
The model that we present in this paper is a first step towards a joint label and language generation formulation for SRL, using the sequence-to-sequence architecture as a starting point.

\begin{figure}
\centering
\includegraphics[width=0.5\textwidth]{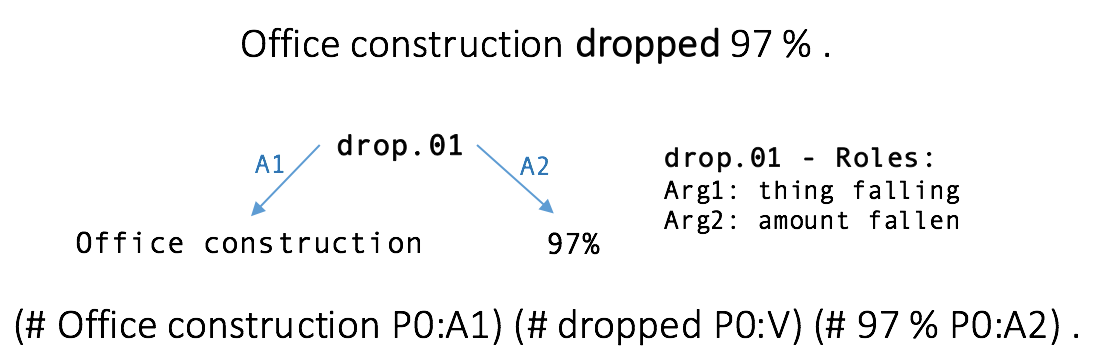}
\caption{\label{fig:propLinear} 
An input sentence (top), its PropBank predicate-argument structure (middle) and its linearized labeled sequence produced by our system.}
\end{figure}

We explore a sequence-to-sequence formulation of SRL that we apply, as a first step, in a classical monolingual setting on PropBank data, as illustrated in Figure \ref{fig:propLinear}. This constrained monolingual setting will allow us to analyze the suitablility of a sequence-to-sequence architecture for SRL, by benchmarking the system performance against existing sequence labeling models for SRL on well known labeled evaluation data.

Sequence-to-sequence (seq2seq) models were pioneered by \citet{Sutskever2014SequenceNetworks}, and later enhanced with an attention mechanism \cite{Bahdanau2014NeuralTranslate,Luong2015EffectiveTranslation}. They have been successfully applied in many related structure prediction tasks such as syntactic parsing \cite{Vinyals2014GrammarLanguage}, parsing into Abstract Meaning Representation \cite{Konstas2017NeuralGeneration}, semantic parsing \cite{DongLapataLogicalForm}, 
and cross-lingual Open Information Extraction \cite{OpenIE}.

When applying a seq2seq model with attention in a monolingual SRL labeling setup, we need to restrict the decoder to reproduce the original input sentence, while in addition inserting PropBank labels into the target sequence in the decoding process (see Figure \ref{fig:propLinear}). To achieve this, we encode each input sentence into a suitable representation that will be used by the decoder to regenerate word tokens as given in the source sentence and introducing SRL labels in appropriate positions to label argument spans with semantic roles. In order to avoid lexical deviations in the output string, we add a copying mechanism \cite{Gu2016IncorporatingLearning} to the model. This technique was originally proposed to deal with rare words by copying them directly from the source when appropriate. We apply this mechanism in a novel way, with the aim of guiding the decoder to reproduce the input as closely as possible, while otherwise giving it the option of generating role labels in appropriate positions in the target sequence.

Our main contributions in this work are:

(i) We propose a novel neural architecture for SRL using a seq2seq model enhanced with attention and copying mechanisms.

(ii) We evaluate this model in a monolingual setting, performing PropBank-style SRL on standard English datasets, to assess the suitability of this model type for the SRL labeling task.

(iii) We compare the performance of our model to state-of-the-art sequence labeling models, including detailed (also comparative) error analysis.

(iv) We show that the seq2seq model is suited for the task, but still lags behind sequence labeling systems that include higher-level constraints.
\section{Model}

\begin{figure*}
  \includegraphics[width=\textwidth,height=6.5cm]{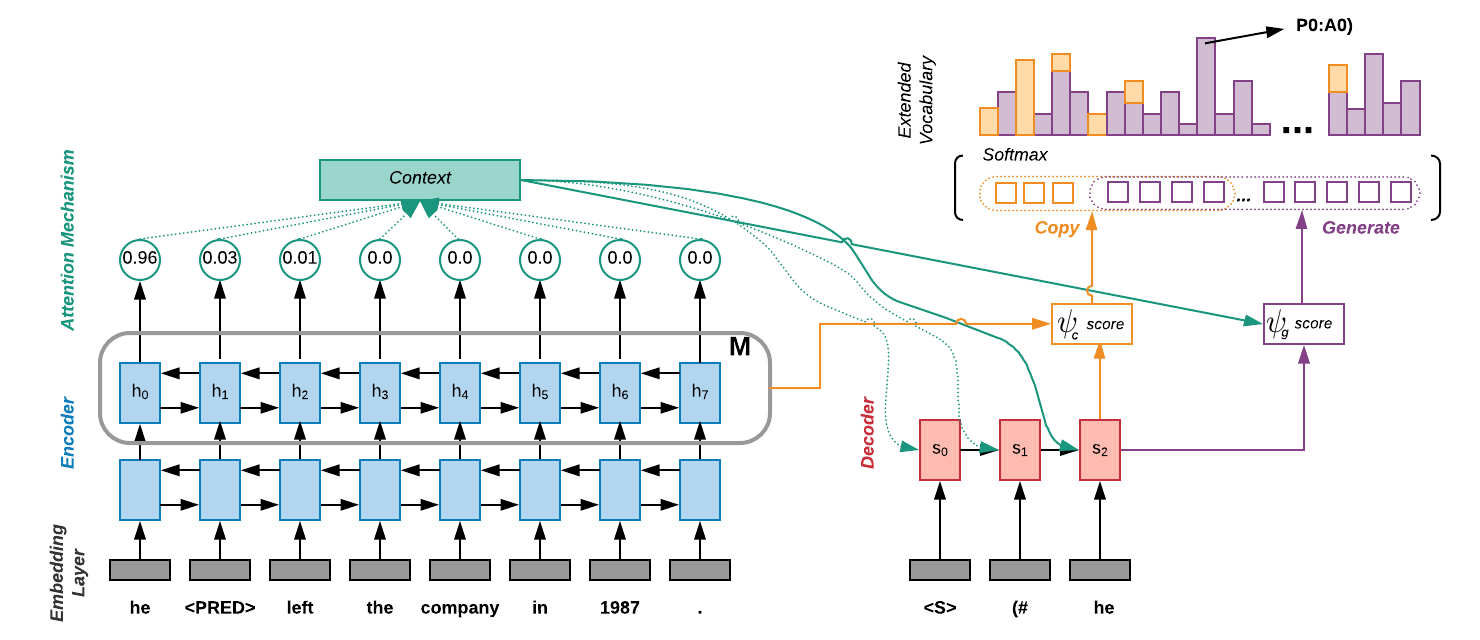}
  \caption{A sequence-to-sequence model for SRL. A score for copying and a score for generating tokens is computed at each time step and a joint softmax determines the probability of the next token over the extended vocabulary of words $\mathcal{V}$, labels $\mathcal{L}$ and current instance words $\mathcal{X}$.}
\label{fig:model}
\end{figure*}

We propose an extension to the Sequence-to-Sequence model of \cite{Bahdanau2014NeuralTranslate} to perform SRL.\footnote{In this work we restrict ourselves to argument labeling.} The model will learn to map an unlabeled source sequence of words $(x_{1}...x_{T_{x}})$ into a  target sequence $(y_{1}...y_{T_{y}})$ consisting of word tokens and SRL label tokens (see Figure \ref{fig:model}). The source sentence, represented as a sequence of dense word vectors, is fed to an LSTM encoder to produce a series of hidden states that represent the input. This information is used by the decoder to recursively generate tokens step-by-step, conditioned on the previous generated tokens and the source by attending the encoder's hidden states as proposed in \citet{Bahdanau2014NeuralTranslate}. On top of this architecture, we add the copying mechanism \cite{Gu2016IncorporatingLearning}, which helps the model to avoid lexical deviations in the output while still having the freedom of generating words and SRL labels based on the context. The attention-based generation and copying mechanism will be competing with each other so that the model learns when to copy directly from the source and when to generate the next token.

In our current setup we restrict role labeling to a single predicate per sentence. If a sentence has more than one predicate, we create a separate copy for each predicate; the same setting was applied in \citet{ZhouEnd-to-endNetworks}. In each sentence copy the predicate whose roles are to be labeled is preceded by a special token $<$\textit{PRED}$>$ that marks the position of the predicate under consideration. This helps the decoder to focus on generating argument labels for that specific predicate (see Table \ref{src_tgts}.)

\subsection{Vocabulary}
We assume a unique vocabulary for both encoder and decoder that comprises the words occurring during training, the out-of-vocabulary token, and the special symbol used to mark the position of the predicate, thus $\mathcal{V}=\left\{v_{1},...,v_{N}\right\}\cup \left\{UNK,<PRED>\right\}$. In addition, we employ a set $\mathcal{L}=\left\{l_{1},...,l_{M}\right\}$ with all the possible labeled brackets 
and a set $\mathcal{X} = \left \{x_{1}...,x_{T_{x}} \right \} $, a per-instance set containing the ${T_{x}}$ words from the current source sequence. Thus, our total vocabulary is defined for each instance as $\mathcal{V}\cup \mathcal{L}\cup \mathcal{X}$. 

The label set $\mathcal{L}$ contains one common opening bracket \textit{(\#} for all argument types to indicate the beginning of an argument span, and several label-specific closing brackets, such as \textit{P0:A1)}, which indicates in this case that the span for argument \textit{A1} is ending (see also Table \ref{src_tgts}).

\subsection{Encoder}
We use a two-layer bi-RNN encoder with LSTM cells \cite{Hochreiter:LSTM} that outputs a series of hidden states $h_{j}=\left [ \overrightarrow{h_j}; \overleftarrow{h_j} \right]$ where each $h_{j}$ contains information about the surrounding context of the word $x_{j}$. We refer to the complete matrix of encoder hidden states as \textbf{M}, since it acts as a memory that the decoder can use to copy words directly from the source.

\subsection{Attention Mechanism} \label{attn}
We use the global dot product attention from \citet{Luong2015EffectiveTranslation} to compute the context vector $c_{i}$:
\begin{equation}
\begin{matrix}
c_{i} = \sum_{j=1}^{T_{x}} \alpha _{ij}h_{j} & \textup{;} & \alpha_{ij}=\frac{\textup{exp}(e_{i,j})} {\sum_{k=1}^{T_{x}}\textup{exp}(e_{i,k})}
\end{matrix} 
\end{equation}
where $e_{i,j}$ is the dot product function between decoder state ${s_{i-1}}$ and each encoder hidden state $h_{j}$.

\begin{table*}[t]
\centering
\resizebox{\textwidth}{!}{%
\resizebox*{!}{\dimexpr\textheight-\lineskip\relax}{%
\centering
\begin{tabular}{lm{19cm}}
\toprule 
Source-1: & The trade figures $<$\textit{PRED}$>$ \textbf{turn out} well , and all those recently unloaded bonds spurt in price .\\
%\midrule
Target-1: & \textit{(\#} The trade figures \textit{P0:A1)} \textbf{ \textit{(\#} turn out \textit{P0:V)}} \textit{(\#} well \textit{P0:A2)} , and all those recently unloaded bonds spurt in price .\\
\midrule
Source-2: & The trade figures turn out well , and all those recently $<$\textit{PRED}$>$ \textbf{unloaded} bonds spurt in price .\\
%\midrule
Target-2: & The trade figures turn out well , and all those \textit{(\#} recently \textit{P0:AM-TMP)} \textbf{ \textit{(\#} unloaded \textit{P0:V)} }\textit{(\#} bonds \textit{P0:A1)} spurt in price .\\
\midrule
Source-3: & The trade figures turn out well , and all those recently unloaded bonds $<$\textit{PRED}$>$ \textbf{spurt} in price .\\
%\midrule
Target-3: & The trade figures turn out well , and \textit{(\#} all those recently unloaded bonds \textit{P0:A1)} \textbf{\textit{(\#} spurt \textit{P0:V)}} \textit{(\#} in price \textit{P0:AM-ADV)} . \\
\bottomrule
\end{tabular}}
}
\caption{A single sentence with three labeled predicates is converted into three different source-target pairs. The symbol $<$\textit{PRED}$>$ in each source marks the predicate for which the model is expected to generate a correct predicate-argument structure.}
\label{src_tgts}
\end{table*}

\subsection{Decoder}
The role of the decoder (a single-layer recurrent unidirectional LSTM) is to emit an output token $y_{t}$ from a learned distribution over the vocabulary at each time step {\it t} given its state $s_{t}$, the previous output token $y_{t-1}$, the attention context vector $c_{t}$, and the memory \textbf{M}. To get this distribution it is necessary to compute two separate modes: one for generating and one for copying.

To obtain the probability of generating $y_{t}$ we use the context vector produced by the attention to learn a score $\psi_{g}$ for each possible token $v_i$ of being the next generated token. We define $\psi_{g}$ as:
\begin{equation}
\begin{matrix}
\psi_{g}(y_{t} = v_{i}) = W_{o}[s_{t};c_{t}], & v_{i} \epsilon \mathcal{V} \cup \mathcal{L}
\end{matrix}
\end{equation}
where $W_{o} \epsilon \mathbb{R}^{N\times 2d_{s}}$ is a learnable parameter and $s_{t}$, $c_{t}$ are the current decoder state and context vector respectively. This means that the model computes a generation score for both words and labels, based on what it is attending on at the current step.

For the probability of copying $y_{t}$ we compute the score $\psi_{c}$ of copying a token directly from the source as:
\begin{equation}
\begin{matrix}
\psi_{c}(y_{t} = x_{j}) = \sigma(h_{j}^{T}W_{c})s_{t}, & x_{j} \epsilon \mathcal{X}
\end{matrix}
\end{equation}
where $W_{c} \epsilon \mathbb{R}^{d_{h}\times d_{s}}$ is a learnable parameter, $h_j$ is the encoder hidden state representing $x_{j}$, $s_{t}$ is the current decoder state, and $\sigma$ is a non-linear transformation; we used {\tt tanh} for our experiments.

Using the two scoring methods, the decoder will have two competing modes: the generation mode, used to generate the most probable subsequent token based on attention; and the copying, used to choose the next token directly from the encoder memory \textbf{M}, which holds both positional and content information of the source. A final mixed distribution is calculated by adding the probability of generating $y_{t}$ and the probability of copying $y_{t}$:
\begin{multline}
p(y_{t} | s_{t},y_{t-1},c_{t},\mathbf{M}) = p(y_{t},\mathbf{g} | s_{t},y_{t-1},c_{t}) + \\ p(y_{t},\mathbf{c}| s_{t},y_{t-1},\mathbf{M})
\end{multline}

We use a {\it softmax} layer to convert the two scores into a joint distribution that represents the mixed likelihood of generating and copying $y_{t}$. Again following \citet{Gu2016IncorporatingLearning}, we define this as:

\begin{equation}
\begin{split}
p(y_{t},\mathbf{g}|\cdot ) &= \left\{\begin{matrix}
\frac{1}{Z}e^{\psi_{g}(y_{t})} &  y_{t}\epsilon \mathcal{V} \cup \mathcal{L} \\ 
0&otherwise 
\end{matrix}\right.
\\
p(y_{t},\mathbf{c}|\cdot ) &= \left\{\begin{matrix}
\frac{1}{Z} \sum_{j:x_{j}=y_{t}} e^{\psi_{c}(x_{j})} &  y_{t}\epsilon \mathcal{X}\\ 
0&otherwise 
\end{matrix}\right.
\end{split}
\end{equation}
where $Z$ is the normalization term shared by the two modes, $Z = \sum_{v \epsilon \mathcal{V}}e^{\psi_{g}(v)} + \sum_{x\epsilon \mathcal{X}}e^{\psi_{c}(x)}$. Since a single {\it softmax} is applied over the copying and generating modes, the network learns by itself when it is proper to copy a word from the source and when it needs to generate a label.

During training, the objective is to minimize the negative log-likelihood of the target token $y_{t}$ for each time-step for both generate mode (given previous generated tokens) and copy mode (given source sequence $X$). We calculate the loss for the whole sequence as:
\begin{equation}
loss = -\frac{1}{T_{y}}\sum_{t=0}^{T_{y}}\log P(y_{t}| y_{<t},X)
\end{equation}

\section{Experimental Setup}

\subsection{Datasets and Evaluation Measures}
We test the performance of our system on the span-based SRL datasets CoNLL-05\footnote{\url{http://www.lsi.upc.edu/~srlconll/home.html}} and CoNLL-12.\footnote{\url{http://conll.cemantix.org/2012/data.html}} These datasets provide the gold predicate as part of the input. Since we focus on argument identification and classification, we provide  this information in the input to the system. We use the standard training, development and test splits and use the official CoNLL-05 evaluation script on both datasets. We compare our results with \citet{CollobertScratch,FitzGerald2015SemanticFactors,ZhouEnd-to-endNetworks} and \citet{He2017DeepNext} who use the same datasets and evaluation script. We show results separately for the Brown and WSJ portion of the CoNLL-05 test dataset.

The CoNLL-05 Shared Task\footnote{\url{http://www.lsi.upc.edu/~srlconll/soft.html}} evaluation script computes precision, recall and F1 measure (the harmonic mean of precision and recall) for the predicted arguments. The script expects prediction-gold pairs that have the same number of words in order to consider them comparable, and only if this is the case, it computes a score. Furthermore, an argument is only considered correct if the words spanning the argument as well as its role label match with gold \cite{CarrerasCoNLL2005}. This means that it is essential to predict perfect argument spans besides the correct role label.

\subsection{Pre-processing}
For our seq2seq model we need to provide sources and targets in a linearized manner. The sequences are sentences with zero or more predicates. Following \citet{ZhouEnd-to-endNetworks}, if a sentence has $n_{p}$ predicates we process the sentence $n_{p}$ times, each one with its corresponding predicate-argument structure. As shown in Table \ref{src_tgts}, we linearize the target side by converting the CoNLL format into sequences of tokens that include brackets indicating the span of the argument and the argument label on the closing bracket. We inform the model about the predicate that it should focus on by adding the special token $<$\textit{PRED}$>$ to the source sequence immediately before the predicate word. This process is entirely reversible and thus we convert the system outputs back to CoNLL format and evaluate the results with the official script. 

\subsection{Training}

Since we process as many copies of sentences as it has predicates, the final amount of sequences is approximately 94K for CoNLL-05 and 185K for CoNLL-12 training sets. We keep linearized sequences up to 100 tokens long and lowercase all tokens. Given this limit, we omit 30 (CoNLL-05) and 900 (CoNLL-12) sequences from training. We initialize the model with pre-trained 100-dimensional GloVe embeddings \cite{penningtonGloVe} and update them during training.\footnote{We also experimented with word2vec word embeddings \cite{Mikolov:w2v} but found GloVe6B (trained on Wikipedia2014+Gigaword5) embeddings to perform better. Available at \url{https://nlp.stanford.edu/projects/glove/}} All the tokens that are not covered by GloVe or that appear less frequently than a given threshold\footnote{We used a threshold of 10 for CoNLL-05 and 15 for CoNLL-12.} in the training dataset are mapped to the $UNK$ embedding. We keep track of this mapping to be able to post-process the sequence and recover the rare tokens. Our vocabulary size is set to $|\mathcal{V}|\approx 20K$ words for CoNLL-05 and $|\mathcal{V}|\approx 18K$ words for CoNLL-12. 

We use Adam optimizer \cite{AdamOpt}, a learning rate $l_{r}=0.001$ and gradient clipping at 5.0. Both encoder and decoder have hidden layer of 512 LSTMs. We use dropout \cite{SrivastavaDropout} of 0.4 and train for 4 epochs with batch size of 6.
\section{Evaluation and Results}

\begin{table}[t]
\centering
\resizebox{0.44\textwidth}{!}{
\begin{tabular}{@{}llllll@{}}
\toprule
               & \multicolumn{3}{l}{CoNLL-05} & \multicolumn{2}{l}{CoNLL-12} \\ 
             & Dev      & WSJ     & Brown   & Dev           & Test         \\
\midrule
\multicolumn{6}{@{}l}{Seq2seq ( attention-only) }  \\
\midrule
same length & 29.19 & 29.98 & 32.24 & - & - \\
brackets & 95.25 & 94.93 & 94.24 & - & - \\
\midrule
\multicolumn{6}{@{}l}{Seq2seq (w/ Attention \& Copying)}  \\
\midrule
same length & 96.71    & 97.15   & 97.24   & 97.46         & 96.07            \\ 
brackets       & 99.91    & 99.82   & 99.88   & 99.97         & 99.93            \\\bottomrule
\end{tabular}
}
\caption{Quality of reproducing words and SRL brackets with seq2seq: Attention-only vs.\ Attention \& Copying, on CoNLL-05 and CoNLL-12 datasets: percentage of correctly reproduced sentence length and percentage of balanced brackets.
}
\label{table_seqs}
\end{table}
Initially, we trained a model using attention only, and it learned to generate balanced brackets (every opening bracket has a corresponding closing bracket within the sequence) without further constraints. Yet, due to its generative nature, many target sequences diverged from the source in both length and token sequences. This was expected, because the system has to learn to generate not only the labels at the correct time-step but also to re-generate the complete sentence accurately. This is a disadvantage compared to the sequence labeling models where the words are already given.

By adding copying mechanism the model successfully regenerates the source sentence in the majority (up to 99\%) of cases, as shown in Table \ref{table_seqs}. Such behavior also enables us to measure the performance of the model as an argument role classifier against the gold standard. Thus, we can benchmark its labeling performance against previous architectures built to solve the SRL task.  

\begin{table}[t!]
\centering
\resizebox{0.48\textwidth}{!}{
\begin{tabular}{@{}lllll|ll@{}} 
\toprule
& \multicolumn{4}{c}{CoNLL-05}& \multicolumn{2}{c}{CoNLL-12}\\
            & dev   & test  & WSJ   & Brown & dev & test\\ \midrule
Collobert & 72.29 & 74.15 & -     & -    & - & - \\
FitzGerald    & 78.3  & -     & 79.4  & 71.2  & 79.2  & 79.6  \\
Zhou \& Xu            & 79.55 & 81.27 & 82.84 & 69.41 & 81.07 & 81.27\\
He           & 81.6  & 81.6  & 83.1  & 72.1  & 81.5  & 81.7 \\
Ours  (min)                 & 76.05  & 76.7 & 78.13  & 66.28 & 73.4 &73.61\\ 
Ours  (max)                 & 77.29  & 77.87 & 79.23  & 68.39 & 75.05 & 75.43\\\bottomrule
\end{tabular}
}
\caption{F1 measure for argument role labeling of our seq2seq model w/ Attention \& Copying on CoNLL-05  and CoNLL-12 dev and test sets, compared to Collobert w/o parser, FitzGerald single model, Zhou \& Xu, and He single model  .
}
\label{SoA-2005}
\end{table}
Table \ref{SoA-2005} displays the overall labeling performance of our copying-enhanced seq2seq model in comparison to previous neural sequence labeling architectures. For sequences that do not fully reproduce the input, we cannot compute appropriate scores against the gold standard. We compute two alternative scores for these cases: \textit{oracle-min}, by setting the score for these sentences to 0.0 F1, and \textit{oracle-max}, by setting their results to the scores we would obtain with perfect (= gold) labels. With these scores, we can better estimate the loss we are experiencing by non-perfectly reproduced sequences (see \ Table \ref{table_seqs}.)

As seen in Table \ref{SoA-2005}, our model achieves an F1 score of 76.05 on the CoNLL-05 development set, and 73.4 on CoNLL-12 (min-oracle), and 77.29 and 75.05 (max-oracle), respectively. While these scores are still low compared to the latest neural SRL architectures, they are above the relatively simple model of \citet{CollobertScratch}. Note also that in contrast to the stronger models of \citet{FitzGerald2015SemanticFactors,ZhouEnd-to-endNetworks} and \citet{He2017DeepNext}, our architecture is very lean and does not (yet) employ structured prediction (e.g. Conditional Random Field), to impose structural constraints on the label assignment. While this is certainly an extension we are going to explore in future work, here we will conduct deeper investigation to learn more about the kind of errors that our unconstrained seq2seq model makes. We report the analysis on CoNLL-05 development set.

\subsection{Analysis}
\textbf{Argument Spans} The model needs to generate labeled brackets at the appropriate time-step, in other words, the prediction of correct spans for arguments. To verify how well it is doing this, we measure how much overlap exists between the generated spans and the gold ones. This is equivalent to computing unlabeled argument assignment. We found that 77.5\% of the spans match the gold spans completely, 21.2\% of spans are partially overlapping with gold spans, and only 1.2\% of the spans do not overlap at all with gold. 

\begin{figure}
\centering
\includegraphics[width=0.4\textwidth]{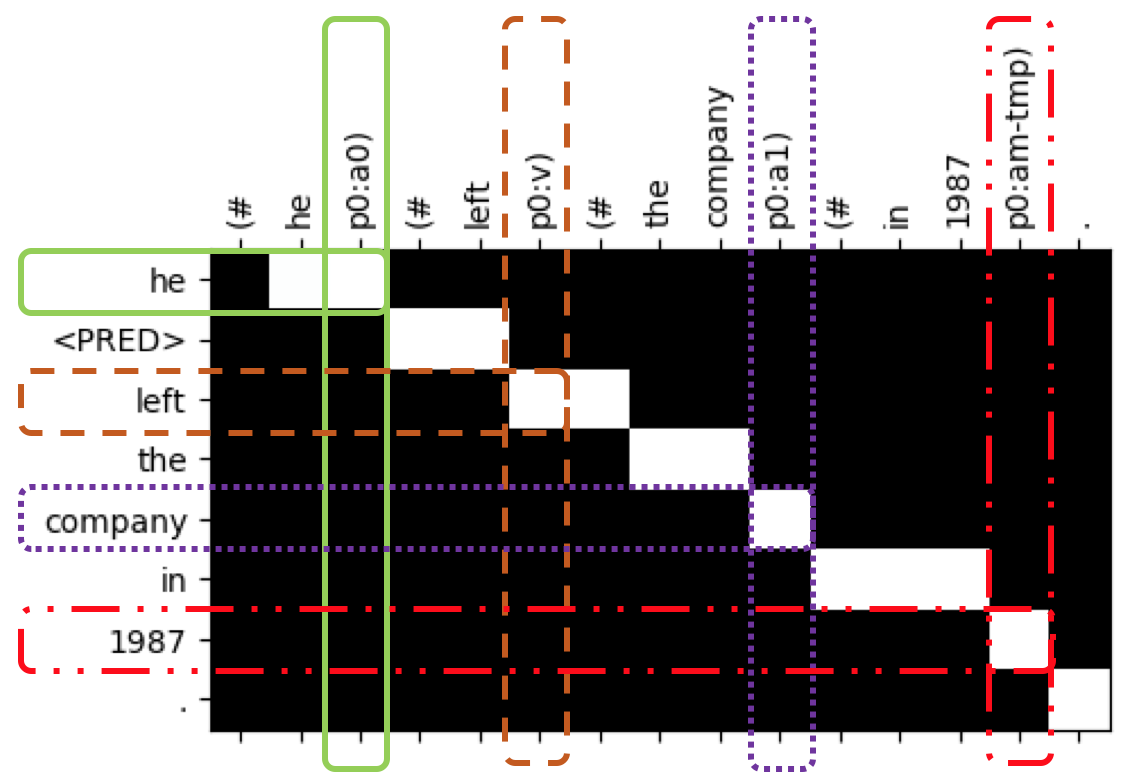}
\caption{\label{fig:attnAlign} 
Example of the alignments learned by the attention mechanism.}
\end{figure}

\begin{figure}
\centering
\includegraphics[width=0.4\textwidth]{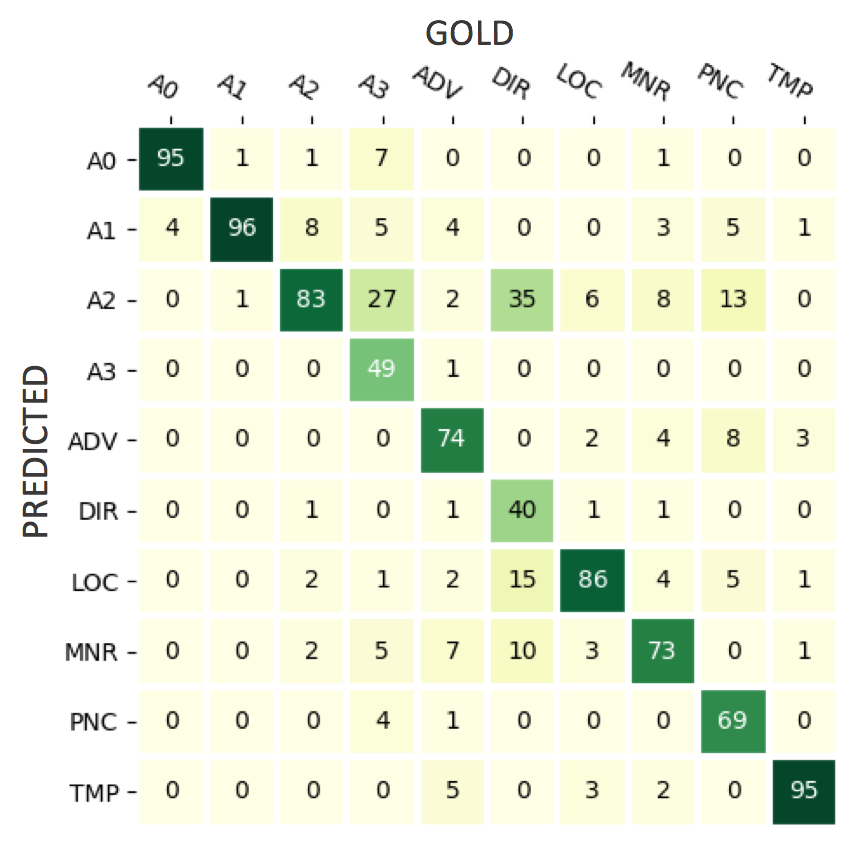}
\caption{\label{fig:confMatrix} 
Confusion matrix showing percentage of predicted labels compared to the gold labels on the CoNLL-05 development set.}
\end{figure}

\textbf{Argument Labels} 
Recall from Section 2 that our model is labeling the sentences as in a translation task. It learns to use information from relevant words in the source sequence, aligning the labels to the argument words via learned attention weights as it is shown in Figure \ref{fig:attnAlign}. This allows us to see where the model is looking when generating the labeled bracket.
The confusion matrix in Figure \ref{fig:confMatrix} shows predicted vs.\ gold labels for all correctly assigned argument spans (i.e., the spans that match the gold boundaries). We observe that the model does very well for A0 and A1 gold roles, and that it causes only few misclassifications for A2. However, it frequently predicts core argument roles A0--A3 for non-argument roles, and also tends to mix predictions among non-core arguments. Since A0 and A1 roles are most frequent in the data, this indicates that the seq2seq model would benefit from more training data, particularly for less frequent roles, to better differentiate roles, and this is more prominent for the ones that are marked with prepositions.

\begin{figure}[t]
\centering
\includegraphics[width=0.48\textwidth]{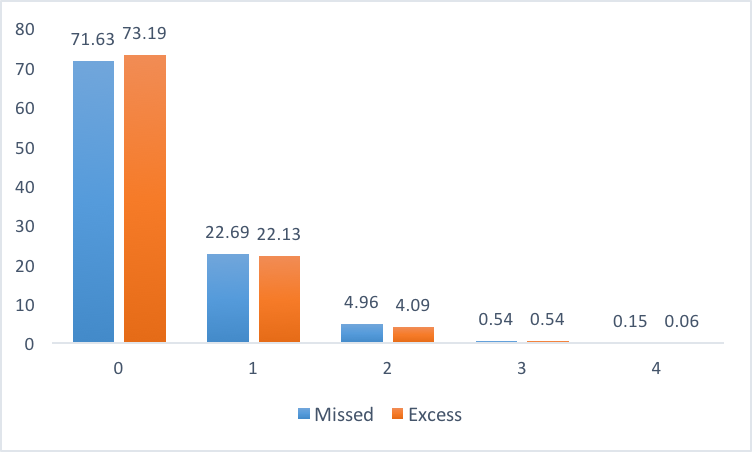}
\caption[Missing and Excess of arguments]{Percentage of sentences with 0,1,2 or more missing (blue) or excess (orange) arguments (seq2seq  w/Copying, CoNLL-05, dev set).}
 \label{fig:excess}
\end{figure}

\textbf{Role co-occurrence and role set constraints}
Despite the absence of more refined decoding constraints, our model learns to avoid generating duplicated argument labels in most of the sequences.  We find duplicated argument labels in less than 1\% of the sequences. Figure \ref{fig:excess} shows that the majority (about 70\%) of sentences do not involve any missing or excess arguments; about 24/20\% of sentences experience a single missing/excess role, and only 5/4\% of the sentences experience a higher amount of missed/excess roles. Overall, missed vs.\ excess arguments are balanced.

\textbf{Sequence Length} Another characteristic of the seq2seq model is that it encodes within a single sequence both words and labeled brackets. This increases the length of the sequences that need to be processed, and it is a well known problem that sequence length affects performance of recurrent neural models, even with the use of attention.

\begin{figure}
\centering
\includegraphics[width=0.5\textwidth]{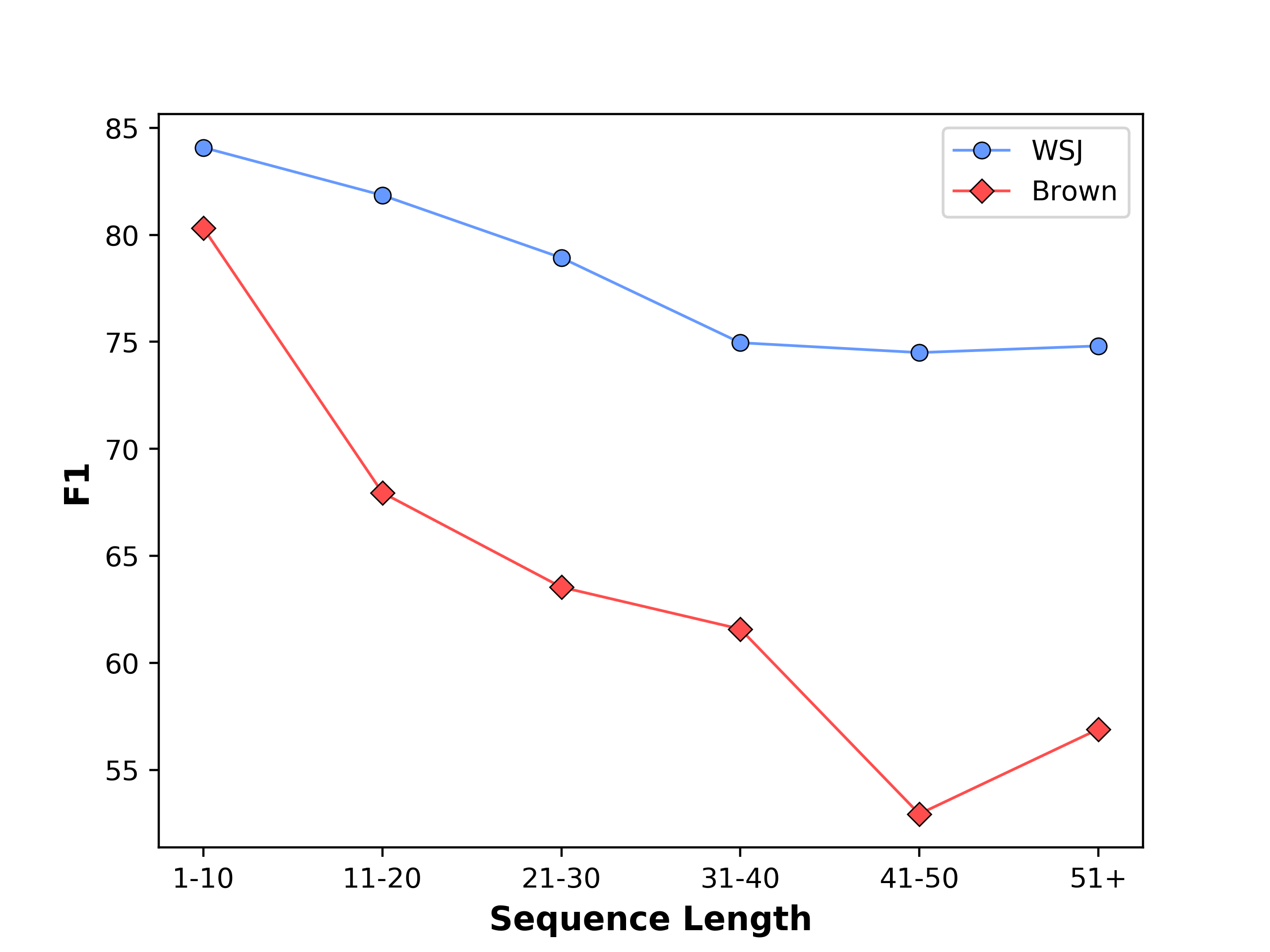}
\caption{\label{fig:seq_len} 
Performance of the model based on the number of tokens that the sequence has.}
\end{figure}

To measure the labeling performance difficulty with increasing sequence length, we partitioned the system outputs in six different bins containing groups of sentences of similar length (see Figure \ref{fig:seq_len}). As expected, the F1 score degrades proportionally to the length of the sequence, especially in sentences with more than 30 tokens.

\begin{figure}[t]
\centering
\includegraphics[width=0.5\textwidth]{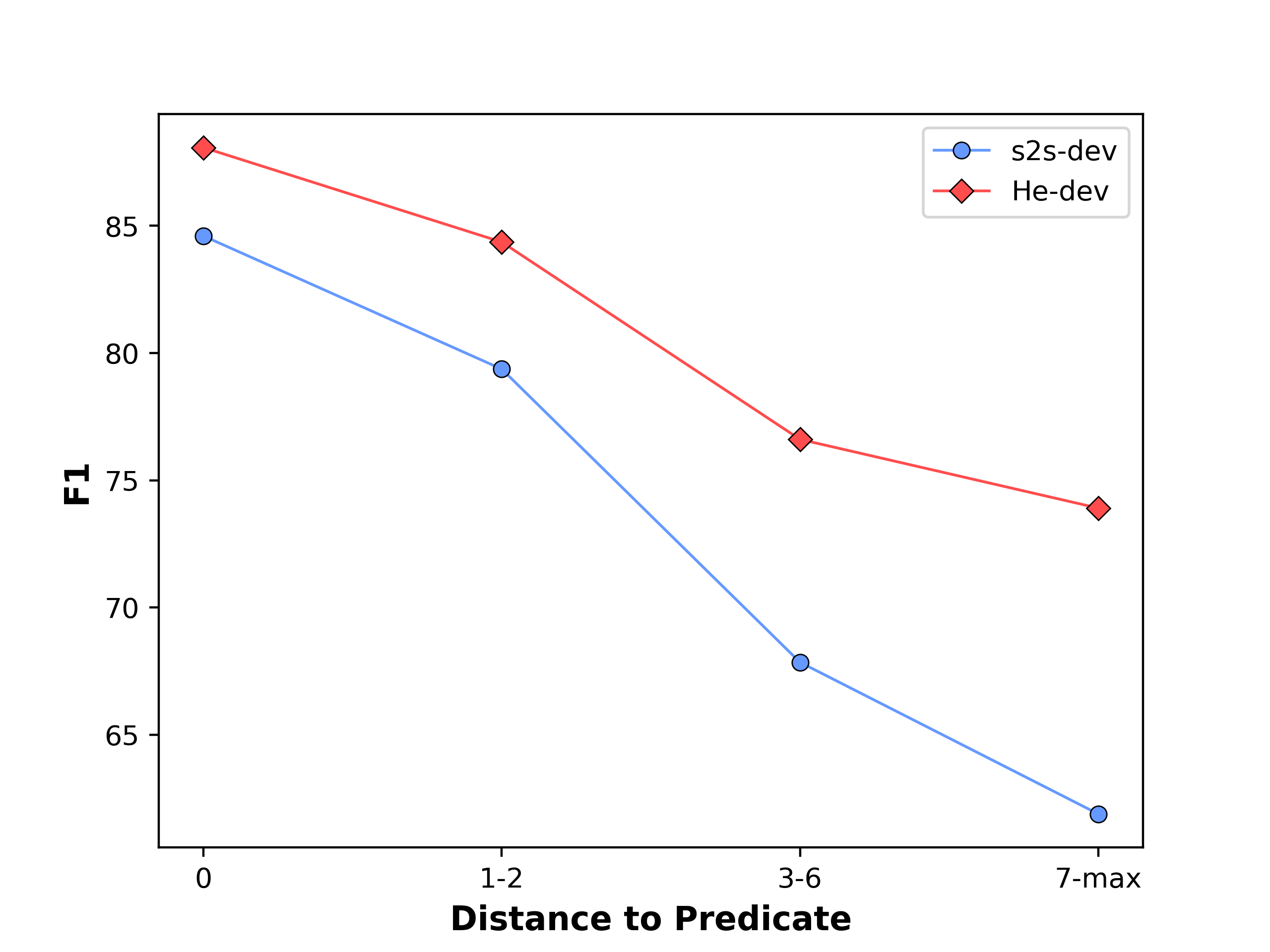}
\caption{
F1 score of arguments in buckets of increasing distances from their predicate, with distance normalized by sentence length (CoNLL-05, dev). We compare our model with \citet{He2017DeepNext}.}
\label{fig:pred_dist} 
\end{figure}

\textbf{Distance to predicate} \citet{He2017DeepNext} show that the surface distance between the argument and the predicate is also 
proportional to the amount of labeling errors. In our model, the distance between argument words and the predicate is even longer because of labeled brackets embedded in the sequence. Figure \ref{fig:pred_dist} displays the F1 score for different token distances between predicate and the respective argument. We see that the seq2seq model follows the same trend as the sequence labeling model, despite the fact that our model has access to the hidden states from the encoded input sentence; however, the real distance between predicate and argument in the decoder is also bigger.

\begin{figure}[t]
\centering
\includegraphics[width=0.5\textwidth]{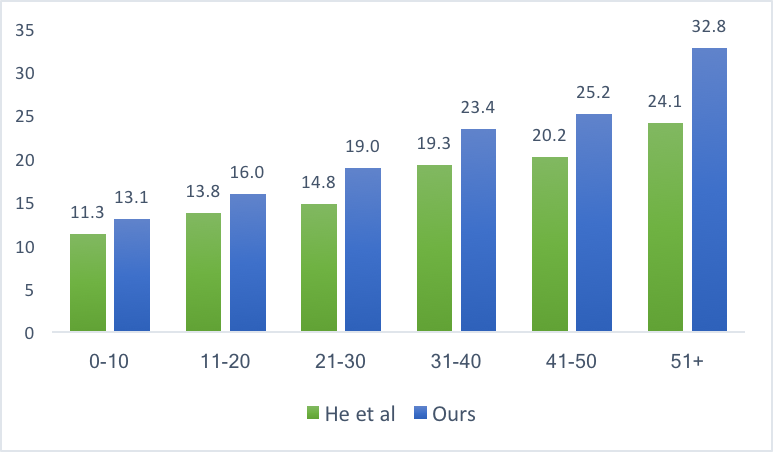}
\caption{
Error ratio of arguments in different regions of the sequences (CoNLL-05, dev).}
\label{fig:err} 
\end{figure}

\textbf{Distance from sentence beginning.} With each token that the model generates in decoding, the distance to the end position of the encoded sentence representation grows. While intuitively we would expect the model performance to degrade with larger distance to the input, it is also true that the model could be more prone to making mistakes at the beginning of the sequence, when the decoder has not yet generated enough context. To investigate this, we traced the ratio of errors that occur in several ranges of the sequence. We can see in Figure \ref{fig:err} that the first intuition was correct, the distance to the encoded representation is proportional to the mistakes that the model makes. We compare the error ratio to \citet{He2017DeepNext} and show that the seq2seq system follows a similar trend but degrades faster with sequence length.

\section{Related Work}

\textbf{Semantic Role Labeling.} Traditional approaches to SRL relied on carefully designed features and expensive techniques to achieve global consistency such as Integer Linear Programming \cite{Punyakanok2008TheLabeling} or dynamic programming \cite{TackstromInferenceSRL}. First neural SRL attempts tried to mix syntactic features with neural network representations. For example, \citet{FitzGerald2015SemanticFactors} created argument and role representations using a feed-forward NN, and used a graphical model to enforce global constraints. \citet{Roth2016NeuralEmbeddings}, on the other hand, proposed a neural classifier using dependency path embeddings to assign semantic labels to syntactic arguments.

\citet{CollobertScratch} proposed the first SRL neural model that did not depend on hand-crafted features and treated the task as an IOB sequence labeling problem. Later, \citet{ZhouEnd-to-endNetworks} proposed a deep bi-directional LSTM model with a CRF layer on top. This model takes only the original text as input and assigns a label to each individual word in the sentence. \citet{He2017DeepNext} also treat SRL as a IOB tagging problem, and use again a deep bi-LSTM incorporating highway connections, recurrent dropout and hard decoding constraints together with an ensemble of experts. This represents the best performing system on two span-based benchmark datasets so far (namely, CoNLL-05 and CoNLL-12). \citet{marcheggiani-frolov-titov:2017} show that it is possible to construct a very accurate dependency-based SRL system without using any kind of explicit syntactic information. In subsequent work, \citet{Marcheggiani2017GCN} combine their LSTM model with a graph convolutional network to encode syntactic information at word level, which improves their LSTM classifier results on the dependency-based benchmark dataset (CoNLL-09). 

\textbf{Sequence-to-sequence models.} Seq2seq models were first discovered as powerful models for Neural Machine Translation \cite{Sutskever2014SequenceNetworks,choSeq2Seq} but soon proved to be useful for any kind of problem that could be represented as a mapping between source and target sequences. \citet{Vinyals2014GrammarLanguage} demonstrate that constituent parsing can be formulated as a seq2seq problem by linearizing the parse tree. They obtain close to state-of-the-art results by using a large automatically parsed dataset. \citet{DongLapataLogicalForm} built a model for a related problem, semantic parsing, by mapping sentences to logical form.  Seq2seq models have also been widely used for language generation (e.g. \citet{karpathy2015,bioGenWiki}) given their ability to produce linguistic variation in the output sequences.

More closely related to SRL is the AMR parsing and generation system proposed by \citet{Konstas2017NeuralGeneration}. This work successfully constructs a two-way mapping: generation of text given AMR representations as well as AMR parsing of natural language sentences. Finally, \citet{OpenIE} went one step further by proposing a cross-lingual end-to-end system that learns to encode natural language (i.e.\ Chinese source sentences) and to decode them into sentences on the target side containing open semantic relations in English, using a parallel corpus for training.

\section{Conclusions}
In this paper we explore the properties of a Sequence-to-Sequence model for identifying and labeling PropBank roles. This is motivated by the fact that using a seq2seq model gives more flexibility for further tasks such as constrained generation and cross-lingual label projection. Another advantage is that our model is a very lean architecture compared to the deep Bi-LSTM of the recent SRL models.

To our knowledge, this is the first attempt to perform SRL using a seq2seq approach. Specific challenges emerged by formulating the problem in this way, such as: (i) the decoding of labels and words within a single sequence; (ii) generating balanced labeled brackets at the correct position; (iii) avoiding repetition of tokens, and especially, (iv) generating labeled sequences that perfectly match the source sentence in order to make the labeled sequence absolutely comparable.

Despite these difficulties, we could show that a sequence-to-sequence model with attention and copying achieves quite respectable labeling performance with a lean architecture and without yet considering structural constraints. For future work we consider extensions towards joint semantic role labeling and constrained generation, to produce new variations of existing labeled data.

\section*{Acknowledgements}
We thank the reviewers for their insightful comments. We are also grateful to \'{E}va M\'{u}jdricza-Maydt for assistance with the PropBank labeling scheme and CoNLL datasets. This research is funded by the Leibniz ScienceCampus Empirical Linguistics
\& Computational Language Modeling, supported by Leibniz Association grant no. SAS2015-IDS-LWC and by the Ministry of Science, Research, and Art of Baden-Wurttemberg.

\bibliography{s2s4srl}
\bibliographystyle{acl_natbib}

\end{document}